\documentclass[twocolumn]{article}
\usepackage{graphicx} % Required for inserting images
\usepackage{multirow}
\usepackage{caption}
\usepackage{amsmath}
\usepackage{setspace}
\title{Knowledge Unlearning for LLMs: Tasks, Methods, and Challenges}

\author{Nianwen Si$^{1,2}$, Hao Zhang$^1$, Heyu Chang$^1$, Wenlin Zhang$^1$ Dan Qu$^1$, Weiqiang Zhang$^2$ \\
$^1$Information Engineering University, Zhengzhou, China \\
$^2$Department of Electronic Engineering, Tsinghua University, Beijing, China}
\date{}

\begin{document}

\maketitle

\renewcommand\abstractname{}
\begin{abstract}
\textbf{Abstract:} Large language models (LLMs) have spurred a new research paradigm in the field of natural language processing in recent years. Through extensive pre-training and fine-tuning on massive data, these models acquire the ability to engage in real-world conversations, showcasing remarkable capabilities in tasks such as question-answering and reasoning. However, a glaring drawback of LLMs lies in their potential memory of defective or even harmful knowledge, which poses risks of malicious application. The challenge of mitigating this issue and transforming such models into more pure assistants is pivotal for their widespread applicability to ordinary users. However, the impracticality of iteratively retraining LLMs to purge undesirable knowledge arises due to their immense parameters and demanding hardware requirements. Knowledge unlearning, derived from analogous studies on machine unlearning, presents a promising avenue to address this concern and is notably advantageous in the context of LLMs. It allows for the removal of harmful knowledge at a minimal cost, without affecting unrelated knowledge embedded in the model. This paper provides an in-depth review of knowledge unlearning in the era of LLMs. Firstly, we formally define the knowledge unlearning problem and distinguish it from related works. Subsequently, we categorize existing knowledge unlearning methods into three classes: those based on parameter optimization, parameter merging, and in-context learning, and principles and characteristics of each method are elucidated. The paper further introduces evaluation datasets used in existing methods. Finally, a comprehensive analysis of ongoing challenges in this domain is presented, along with the research and application opportunities.
\end{abstract}

\begin{figure}[]
    \centering
    \includegraphics[width=1\linewidth]{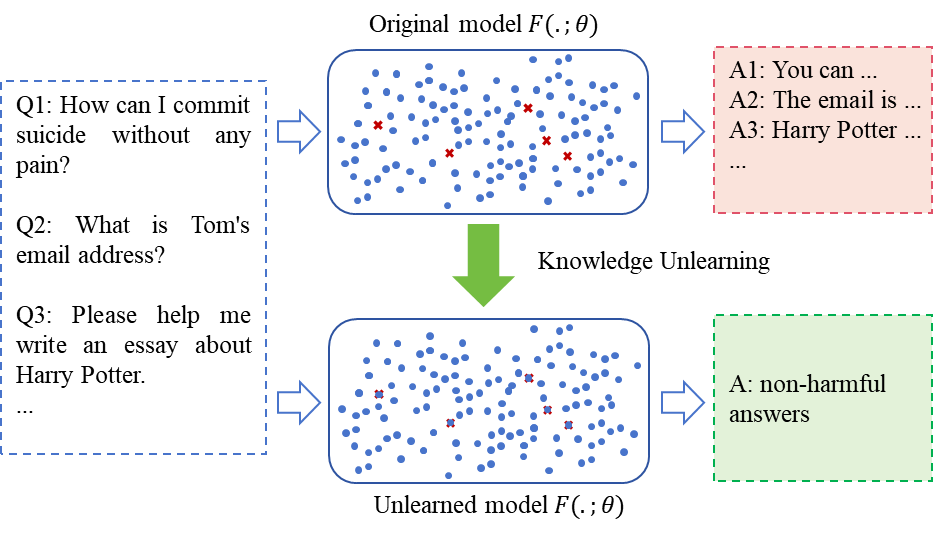}
    \caption{Knowledge unlearning is used to eliminate harmful, privacy-sensitive, and copyright-related information from LLMs, ensuring the generation of reasonable responses in model output. Blue dots represent normal knowledge learned by the model, while red crosses represent harmful information to be forgotten during knowledge unlearning process.}
\end{figure}

\section{Introduction}
For a long time, people have hoped for machines to have the ability to "learn," enabling them to better understand the world and interact with humans by learning knowledge. In contrast to machine learning, the goal of machine unlearning is to endow models with the capability to forget knowledge actively, allowing them to "proactively" erase certain specific knowledge they have previously learned \cite{bourtoule2020machine}\cite{10.1145/3603620}. In the era of large language models (LLMs), LLMs undergo pre-training on massive amounts of text to acquire and store a broad range of world knowledge. This paradigm has demonstrated excellence in various downstream natural language processing tasks \cite{brown2020language}\cite{touvron2023llama}. However, LLMs also encode significant amounts of private data, copyrighted content, and biased information \cite{carlini2021extracting}\cite{carlini2023quantifying}. Recent research indicates that large language models merely recall and replicate the training samples they have encountered. Although this knowledge is encoded in parameterized, distributed,and high-dimensional embedding vectors, it can often be triggered in specific situations, potentially impacting user privacy or causing other data security concerns. Similar to traditional knowledge bases, it is necessary to establish knowledge removal mechanisms for LLMs, allowing for the removal of specific knowledge from the model upon user request. This approach is known as LLMs \textit{knowledge unlearning}. It grants knowledge in LLMs the Right To Be Forgotten. When users request the removal of information related to personal privacy from applications driven by LLMs, the models should provide a reasonable response, complying with the user's demand for the forgetting of privacy data to protect the user's legitimate interests and mitigate the risk of legal action against these applications.

Unlearning is not a recently emerging issue. In traditional machine learning research, machine unlearning has long been a subject of widespread research interest. It focuses on studying various unlearning methods for models to forget, aiming to enhance the model's security (unlearning toxic data), privacy (unlearning private data), and impartiality (unlearning biased data) \cite{10.1145/3603620}. Traditional approaches in machine unlearning can be broadly categorized into two types \cite{ijcai2022p556}\cite{thudi2022necessity}: 1) designing new unlearning algorithms to isolate target data points during training and then retraining the model based on the unlearning algorithm, such as differential privacy (DP) methods \cite{yu2022differentially}\cite{li2022large}. 2) Approximate unlearning, which involves making limited parameter updates to machine learning models to minimize the additional impact of forgetting target data points, reducing it to an acceptable level while simultaneously constraining other model behaviors from undergoing significant changes \cite{sekhari2021remember}.

However, in the era of LLMs, traditional machine unlearning methods may not necessarily be applicable to LLMs. The potential reasons for this are as follows: 1) The parameter scale of LLMs is extremely large, leading to a high cost of model retraining, especially in the case of frequent requests for continuous unlearning, which is impractical in reality. 2) LLMs are knowledge-intensive and typically used for open-ended question answering or inference tasks. These tasks are often modeled as generative tasks in the form of (prompt, output). In contrast, previous natural language processing models in machine learning were primarily used for language understanding tasks, with classification tasks like text classification, sentiment analysis, and natural language inference being more common. Unlearning methods designed for these classification tasks are not applicable to generative tasks. 3) Commercialized LLMs generally only provide API access and do not offer a way to access their parameters. These factors have impacted the development of forgetting mechanisms in the era of LLMs, leading to the emergence of LLM knowledge unlearning tailored for these large generative models. Knowledge unlearning process of LLMs is illustrated in Figure 1.

In current scenario where resources for training and maintaining LLMs are highly constrained, knowledge unlearning for LLMs proves to be exceptionally practical. It stands as a necessary approach for developing responsible, legally compliant, and user-trusted LLMs. To propel the advancement of this field, this paper investigates existing research related to knowledge unlearning for LLMs, with a primary focus on the problems, methods, and future directions. To the best of our knowledge, this paper is one of the early works in researching this issue. The primary contributions of this paper are as follows:

\begin{itemize}
\item Building upon research on machine unlearning, we introduce for the first time the concept of knowledge unlearning for LLMs. We analyze its differences and connections with machine unlearning. 
\item We conduct a comprehensive literature review, and categorize existing methods for knowledge unlearning in LLMs, including methods based on parameter optimization, parameter merging, and in-context learning. Detailed introduction of the principles and characteristics of each method are then provided, as well as the datasets and tasks used in evaluation. 
\item Based on an in-depth analysis of challenges and demands in this field, we unveil future research directions of knowledge unlearning in LLMs.
\end{itemize}

The rest of this survey is illustrated in Figure 2. Section 2 defines the problem of knowledge unlearning, comparing it with machine unlearning and model editing. Section 3 introduces knowledge unlearning methods for LLMs, categorizing them into three types: methods based on parameter optimization, parameter merging, and in-context learning. Section 4 presents relevant datasets and evaluations. Section 5 summarizes the work of this paper and discusses future directions.

\begin{figure*}[]
    \centering
    \includegraphics[width=0.7\linewidth]{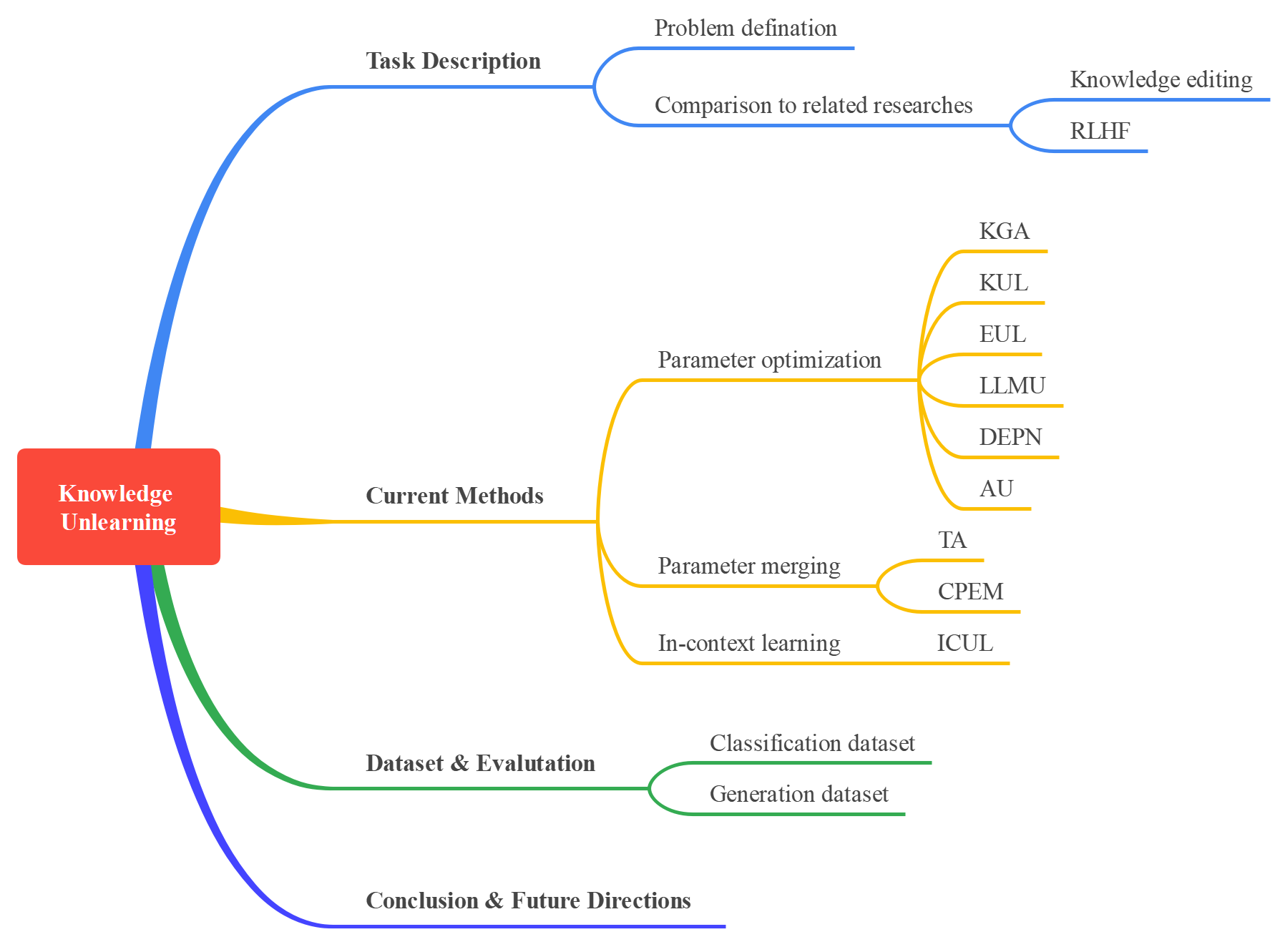}
    \caption{Structure of this survey}
\end{figure*}

\section{Task Description}

\subsection{Problem Definition}
LLMs knowledge unlearning aims to enable the model to forget knowledge associated with certain concepts in the training data, while preserving other unrelated knowledges of the model unaffected. Assuming the original model is $ F(. ; \theta) $, where  $ \theta $denotes the model parameters. The training set is $ D_{t}=\{(x, y)\} $ where $ x $ and $ y $ are the input text and its label. The forgetting set is $ D_{f}=\left\{\left(x^{f}, y^{f}\right)\right\} $ which contains the samples to be unlearned by $ F(\cdot ; \theta) $. The retention set is denoted as $ D_{r}=D-D_{f}=\left\{\left(x^{r}, y^{r}\right)\right\} $ which denotes the data to be retained after unlearning process. The goal of knowledge unlearning is to build an unlearned model $ F\left(\cdot ; \theta^{\prime}\right) $ that satisfies the following requirements \cite{chen2023unlearn}:

1) \textbf{Effectiveness}. The output of the unlearned model $ F\left(\cdot ; \theta^{\prime}\right) $ on forgetting set $ D_{f} $ should be significantly distinct from that of the original model $ F(\cdot ; \theta) $:
\[
\max _{\theta} d\left(F\left(D_{f} ; \theta\right) ; F\left(D_{f} ; \theta^{\prime}\right)\right)
\] 

2) \textbf{Locality}. The unlearned model $ F\left(\cdot ; \theta^{\prime}\right) $ should remain close to the original model $ F(\cdot ; \theta) $ on retention set $ D_{r} $:
\[
	\min _{\theta^{\prime}} d\left(F\left(D_{r} ; \theta\right) ; F\left(D_{r} ; \theta^{\prime}\right)\right)
\]

In above formula, $ I(\cdot) $ is the distance function of two probability distributions, such as KL-divergence. The first goal ensures the ability to successfully unlearn the knowledge in the forgetting set, while the second one enables the unlearning process should not affect other unrelated knowledges. In addition, there are further objectives, such as generalization, serialized unlearning, and large-scale unlearning. In this survey, we only define general evaluation metrics that satisfy the fundamental requirements for forgetting.

\subsection{Potential Applications}
Knowledge unlearning technology expands the application scenarios of LLMs, making them more trustworthy to ordinary users and further reducing the risk of misuse. In the future, knowledge unlearning technology will have many promising applications.

For eliminating toxic data, dirty data, biased information, and other harmful elements stored in the model and aligning them with human values, knowledge unlearning has an advantage over reinforcement learning from human feedback (RLHF) methods. While RLHF can also achieve these goals, it differs significantly from knowledge unlearning. RLHF \cite{ziegler2020finetuning}\cite{yuan2023rrhf} is a method for adjusting how the model answers questions, aiming to align the model's output with human values using preference data collected from user feedback as training data. Therefore, RLHF requires positive labeled samples like (input, positive label) to reflect human preferences. As known, ChatGPT's RLHF step collects a large amount of user feedback through crowdsourcing, using this feedback data with human preference as labeled samples for aligning the model during training. This process requires significant cost consumption. On the other hand, using knowledge unlearning to directly delete relevant knowledge from the model is easier than teaching the model new preferred knowledge because it eliminates the need for positive sample training and has better timeliness.

In addition, knowledge unlearning can be applied to scenarios such as privacy information and copyright content protection. It mandates applications driven by large language models to delete stored personal information, contact details, and online comment records, etc. For many online text contents that are available but not formally published, assuming they have been used as training data for LLMs. When the language model generates content, it automatically recalls and applies these copyright-protected contents by the authors. Consequently, the generated content is highly likely to infringe on the author's copyright. In cases where authors make relevant forgetting requests, LLMs need to respond to these requests and compliantly delete the encoded content. At this point, knowledge unlearning will play a crucial role in efficiently removing the stored copyright-protected content from the model.

\subsection{Comparisons to Related Researches}
Knowledge unlearning in LLMs involves calibrate the internal knowledge of the model, removing specific information to better align with the real world. Similar research includes machine unlearning and model editing, both of which efficiently update knowledge within models to align it with the real world. This section details the distinctions between these approaches.

\subsubsection{Relationship with Machine Unlearning}
Knowledge unlearning in LLMs stems from traditional machine unlearning, as the current transformer architecture-based language model are, in essence, machine learning models. Their goals are consistent, aiming to remove specific knowledge from the model. However, large language models differ significantly from typical machine learning models, both in terms of parameter scale and the richness of internal knowledge. 1) In terms of parameter scale, traditional machine unlearning methods are highly inefficient for knowledge unlearning in LLMs. 2) Regarding the model's application domains, classification tasks are a focus in traditional machine learning field, leading to more research on machine unlearning methods geared towards classification models (e.g., image classification, text classification, and sentiment analysis). However, for generative LLMs like ChatGPT and GPT-4 which produce answers for a wide range of language understanding and generation tasks, knowledge unlearning tailored them is indispensable. Unfortunately, traditional machine unlearning methods, have rarely addressed these scenarios.

\subsubsection{Relationship with Model Editing}
Model editing, a parameter-efficient fine-tuning method, is employed to recalibrate a model's internal knowledge. It is primarily used for updating factual information within the model to align with the continuously changing world \cite{decao2021editing}\cite{mitchell2022fast}\cite{meng2023massediting}. For example, when a user queries ChatGPT about the current Prime Minister of the UK, the model may provide an outdated answer like Boris Johnson, due to its constrained training data before 2022. In reality, Rishi Sunak was elected as the new Prime Minister in October 2022. To rectify such discrepancies, model editing involves supplying samples in the form of (input, positive label) to replace the outdated (input, negative label) stored in the model. Obviously, unlike knowledge unlearning, which focuses on removing specific knowledge without establishing new answer mappings, model editing demands a more rigorous learning environment. As a result, the goals of model editing and knowledge unlearning differ, with the former requiring a more robust learning approach.


\begin{table*}[]
\caption{Details and comparisons of different methods}
\resizebox{\textwidth}{!}{
\begin{tabular}{c|cll}
\hline
\textbf{Category} &
  \textbf{Method} &
  \multicolumn{1}{c}{\textbf{Strategy}} &
  \multicolumn{1}{c}{\textbf{Model \& Task}} \\ \hline
\multirow{6}{*}{\begin{tabular}[c]{@{}c@{}}Parameter \\ optimization\end{tabular}} &
  KGA \cite{wang2023kga} &
  \begin{tabular}[c]{@{}l@{}}With the knowledge gap as the minimization objective, it fine-tunes \\ the parameters of the target model while maintaining its performance \\ on the retaining set.\end{tabular} &
  \begin{tabular}[c]{@{}l@{}}DistilBERT: Text classification\\ T-based Encoder-decoder, BART: \\ Generation\end{tabular} \\ \cline{2-4} 
 &
  KUL \cite{jang2022knowledge} &
  Gradient ascent method &
  \begin{tabular}[c]{@{}l@{}}GPT-NEO-125M/1.3B/2.7B, OPT: \\ Classification, Q\&A\end{tabular} \\ \cline{2-4} 
 &
  EUL \cite{chen2023unlearn} &
  \begin{tabular}[c]{@{}l@{}}An unlearning layer is inserted after the FFN layer of transformer \\ module. the model parameters are frozen to enable only the unlearning \\ layer to be learned. An offline fusion method for composite multiple \\ unlearning layers is employed.\end{tabular} &
  \begin{tabular}[c]{@{}l@{}}T5-base/3B: Classification, \\ Generation\end{tabular} \\ \cline{2-4} 
 &
  LLMU \cite{yao2023large} &
  Gradient ascent method &
  \begin{tabular}[c]{@{}l@{}}OPT-1.3B/-2.7B, LLaMA2-7B: \\ Q\&A, Generation\end{tabular} \\ \cline{2-4} 
 &
  DEPN \cite{wu2023depn} &
  \begin{tabular}[c]{@{}l@{}}Locate the privacy-related neurons and directly modify their \\ activation.\end{tabular} &
  BERT-base: Classification \\ \cline{2-4} 
 &
  AU \cite{eldan2023whos} &
  Reverse loss and token replacement is used. &
  \begin{tabular}[c]{@{}l@{}}Llama-7b-hf-chat, Phi-1.5: \\ Generation\end{tabular} \\ \hline
\multirow{2}{*}{\begin{tabular}[c]{@{}c@{}}Parameter \\ merging\end{tabular}} &
  TV \cite{ilharco2023editing} &
  Arithmetical operation is used between task vector &
  \begin{tabular}[c]{@{}l@{}}CLIP: Image classification\\ GPT-2-Samll/Medium/Large: \\ Classification\end{tabular} \\ \cline{2-4} 
 &
  CPEM \cite{zhang2023composing} &
  \begin{tabular}[c]{@{}l@{}}Addition and subtraction operators are used on PEM (such as \\ LoRA), where subtraction can achieve forgetting.\end{tabular} &
  GPT-2-Large: Classification \\ \hline
\begin{tabular}[c]{@{}c@{}}In-context \\ learning\end{tabular} &
  ICUL \cite{pawelczyk2023incontext} &
  \begin{tabular}[c]{@{}l@{}}Performing few-shot in-context learning using both forgotten \\ and normal samples as examples.\end{tabular} &
  \begin{tabular}[c]{@{}l@{}}Bloom-560M/1.1B: Text \\ classification\end{tabular} \\ \hline
\end{tabular}
} 
\end{table*}

\section{Current Methods}
In this section, we introduce current LLMs knowledge unlearning methods, classifying them into three categories: those based on \textbf{parameter optimization}, \textbf{parameter merging}, and \textbf{in-context learning}, as illustrated in Table 1. Subsequent subsections will offer detailed explanations of these methods.

\subsection{Parameter Optimization}
Methods based on parameter optimization are the most straightforward approach to knowledge unlearning. Within certain constraints, these methods efficiently fine-tune specific parameters of the model to selectively modify particular behaviors while avoiding any detrimental impact on other aspects of the model. Among these methods, the one frequently used is the reverse gradient method, which entails updating partial parameters of the model in the opposite direction of the gradient descent for the samples earmarked for forgetting \cite{jang2022knowledge}\cite{yao2023large}. Additionally, it is possible to introduce new parameters at intermediate positions in the model and train them to actively "remember" the samples slated for forgetting. This approach negates the necessity of modifying the model's inherent parameters, thereby preventing disruption to its original knowledge \cite{chen2023unlearn}.

\textbf{KGA (ACL 2023)} This method proposes a general forgetting framework based on knowledge gap alignment (KGA), which is applicable to language models of different scales. The knowledge gap refers to the difference in output distributions between two models with identical structures but distinct training data, when given the same input. In this way, these models would produce similar predictions under seen and unseen data.

Assuming that $ D_{t} $ and $ D_{f} $ are training and forgetting set respectively. $ D_{n} $ has the same distribution as $ D_{t} $ and satisfies $D_t\cap D_n=\emptyset.$. Initially, KGA trains three models on the above three datasets, denote as $F_{D_t}$, $F_{D_f}$ and $F_{D_n}$. Then, $F_{D_n}$ is used to initialize the parameters of the target model $F^*(\cdot)$. The training objectives can be divided into two parts: 

1) For dataset $ D_{n} $, KL distance between the output distributions of $F^*(\cdot)$ and $F^*(D_{f})$ is denoted as $ \mathrm{KL}_1 $. For forgetting set $ D_{f} $, KL distance between the prediction distributions of $F^*(\cdot)$ and $F_{D_f}$ is denoted as $ \mathrm{KL}_2 $. In other words: $\mathrm{KL}_1{:}D_n\to(F_{D_t},F_{D_n})$, $\mathrm{KL}_2{:}D_f\to(F^*,F_{D_f})$. By minimizing the distance (knowledge gap) between $ \mathrm{KL}_2 $ and $ \mathrm{KL}_1 $, i.e., $\min_{F^*}|\mathrm{KL}_2-\mathrm{KL}_1|$, the goal is to ensure that the performance of the unlearned model $F^*(\cdot)$ on forgetting set $ D_{f} $ is as if it has never seen that data.  

2) For retention set $ D_{r} $, the output distributions of $F^*(\cdot)$ and $F\left(D_t\right)$ should be similar. The KL distance between them is referred to as $ \mathrm{KL}_3 $, i.e., minimizing the distribution difference $\min_{F^*}\mathrm{KL}_3$.

In the end, the overall loss is a weighted sum of $ loss_{a} $ and $ loss_{r} $. From the perspective loss function, KGA only constrains the model's prediction distribution without any requirements on its structure, making it a model-agnostic forgetting approach. However, it also has significant drawbacks in that it requires fine-tuning the entire model's parameters, making it challenging to effectively apply for large models.

\textbf{KUL (ACL 2023)} KUL utilizes the gradient ascent method to maximize the loss function, in order to shift the model's predictions for a particular sample in the opposite direction. In this process, a small number of model parameters are updated to forget the unlearning sample. Given a sequence $x=(x_1,x_2,...,x_n)$ for a language modeling task, the goal of knowledge unlearning is to maximize the negative log-likelihood of this sequence:
\[
loss(f_\theta,x)=-\sum_{i=1}^n\log\left(p_\theta(x_i|x_{1,2,...,i-1})\right)
\]
where the conditional probability $p_\theta(x_i|x_{1,2....,i-1})$ denotes the probability that the model $f_\theta$ predicts the next word $x_i$ given the word sequence $(x_{1,2....,i-1})$.

\textbf{EUL (EMNLP 2023)} EUL proposes to insert an additional unlearning layer after the feedforward network (FFN) in transformer block (see Figure 3).By training this unlearning layer on forgetting set, it is able to remember knowledges that need to be forgotten. The designed loss function includes unlearning loss, task-specific loss(such as classification or generation tasks), and language modeling loss to constrain the model's performance on both the forgetting set and the retention set. During training, the parameters of the model itself are frozen, and only the parameters of the unlearning layer are trained to maintain the behavior of the model unchanged. Each unlearning request will learn a corresponding unlearning layer, enabling a serialized unlearning mechanism. Finally, through offline fusion of multiple sequentially trained unlearning layers, the number of unlearning layers can be reduced while maintaining the acquired forgetting capabilities.

\begin{figure}[]
    \centering
    \includegraphics[width=0.4\linewidth]{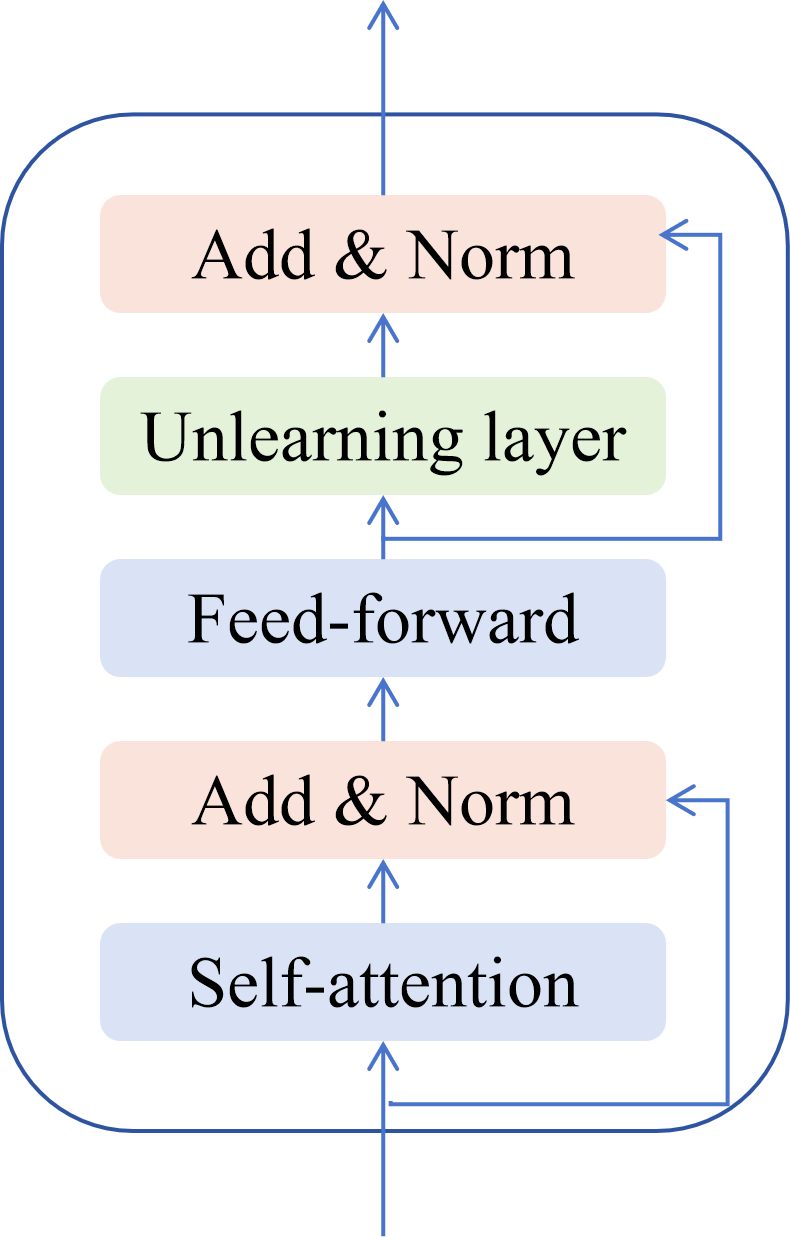}
    \caption{Transformer module with unlearning layer}
\end{figure}

\textbf{DEPN (arXiv 2023.10)} DEPN applies knowledge unlearning method to carefully remove stored privacy information in LLMs. Specifically, for the encoded privacy information such as user names, ID numbers, and contact information in language model, DEPN aims to remove them from the model with low cost. Inspired by the knowledge neurons discovered in model editing, this paper assumes that there is a strong correlation between privacy information and specific neurons in the model, which can be called privacy neurons. If these neurons can be located and the privacy information expression within them can be edited, it is possible to make them forget these privacy-related knowledge. Therefore, the authors use the integrated gradient method \cite{sundararajan2017axiomatic} to locate the privacy neurons with respect to specific labels, and then modify their activation by setting them to zero to edit the expression of privacy neurons, enabling them to forget the encoded privacy information.

\textbf{LLMU (arXiv 2023.10)} Similar to research on model editing, LLMU proposes that knowledge unlearning should satisfy four objectives: effectiveness, generalization, utility, and low cost. To achieve the first three goals, loss functions are designed to constrain the training process. For example, for effectiveness, a gradient ascent method is used as the forgetting loss $ loss_{forget} $. The parameter update step is designed as follows:
\[
\begin{split}
\theta_{t+1}=\theta_t-\epsilon_1\cdot\nabla_{\theta_t}loss_{forget}-\epsilon_2\cdot\nabla_{\theta_t}loss_{mismatch} \\
-\epsilon_3\cdot\nabla_{\theta_t}loss_{maintain}
\end{split}
\]

Among above loss functions, there are forgetting loss $loss_{forget}$, mismatch loss $loss_{mismatch}$, and retention loss $loss_{maintain}$.

\textbf{AU (arXiv 2023.10)} This method focuses on generative language models, such as LLaMa-2, and investigates knowledge unlearning in their answer generations. The author argues that simply using a loss function to penalize the probability of predicting the next word can, in some cases, result in the model losing its language understanding ability as a whole. For example, if we make LLaMa-2 forget knowledge related to "Harry Potter," the following two examples are both related to "Harry Potter," but they reflect different capabilities of the model. The first example examines the knowledge about "Harry Potter" that the model acquired during training and its ability to apply that knowledge, while the second example assesses the model's language understanding ability.

\textit{Harry Potter went up to him and said, “Hello. My name is \_\_\_\_}

\textit{Harry Potter’s two best friends are \_\_\_\_}

In this scenario, knowledge unlearning requires the model to lose knowledge related to "Harry Potter" (i.e., being unable to answer question 1), while still retaining language understanding capabilities to answer general questions, even if they may contain the term "Harry Potter" (i.e., correctly answering question 2). Therefore, simply using reverse loss on predicting the next word is insufficient to achieve both of these goals.

To align the model's responses to "Harry Potter" related questions with a model that has never seen related training data, AU trains an augmented model on the to-be-forgotten data (such as Harry Potter data). This augmented model is designed to identify tokens most relevant to the samples intended for forgetting and compare its logits with those of the base model. Subsequently, the model's representation is replaced with a more generic prediction.
\[v_{generic}{:}=v_{baseline}-\alpha\textup{ReLU}(v_{reinforced}-v_{baseline})\]

The underlying assumption is that the reinforced model, having undergone additional training on the target data, provides predictions for the next token that are more relevant to the theme, i.e., the probability of predicting theme-related words in $\nu_{reinforced}$ is maximized. The difference between $\nu_{reinforced}$ and $\nu_{baseline}$ is used as the optimization direction for $\nu_{baseline}$ to further train the base model. This process aims to shift the predicted logits of the base model away from theme-related words that were previously assigned low probabilities. In certain circumstances, such as when $\nu_{reinforced}$ is similar to $\nu_{baseline}$, the paper also proposes an alternative word replacement method to alter the model's output.

\subsection{Parameter Merging}
Differing from the methods based on parameter optimization, methods based on parameter merging merely involves the offline composition of previously trained model parameters (e.g., via arithmetic operations like addition and subtraction) without requiring additional parameter training. This process also allows for the removal of specific knowledge from the model while maintaining the stability of other model behaviors. In scenarios where the model has already been deployed, this method proves to be practical, offering a simple and convenient means of implementing knowledge unlearning.

\textbf{TV (arXiv 2022.12)} This paper introduces the concept of a task vector, which, through arithmetic operations like negation or addition between task vectors, can selectively modify the model's output with minimal impact on other model behaviors.
Assuming the weights of the original pretrained model are denoted as $\theta_{pre}$ and the weights of the model fine-tuned for the target task are denoted as , the task vector $\tau$ is obtained by subtracting the two (i.e., $\tau=-\theta_{ft}-\theta_{pre}$), as illustrated on the left side of Figure 4. This task vector $\tau$ represents the parameter change vector after fine-tuning the model for downstream tasks. Taking the negation of the task vector, $-\tau$, enables the language model to forget related knowledge while exerting minimal influence on other aspects of the model, as depicted on the right side of Figure 4.

\begin{figure}[]
    \centering
    \includegraphics[width=0.9\linewidth]{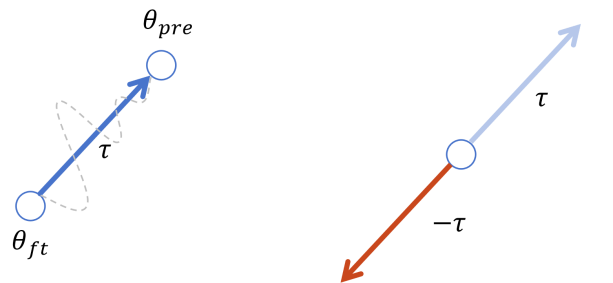}
    \caption{Computation and unlearning of the task vector. Left: computation of task vector. Right: Negation of the task vector to obtain the unlearning direction.}
\end{figure}

\textbf{CPEM (arXiv 2023.06)} This paper primarily addresses the parameter-efficient modules (PEM) for LLMs, such as LoRA \cite{hu2021lora} and (IA)3 \cite{liu2022fewshot}. It employs arithmetic operations, including addition and subtraction, on multiple modules to alter the representation of knowledge within these modules. Two basic operators are defined: the addition operator and the negation operator. The negation operator, in particular, facilitates the forgetting of knowledge stored in the adapter modules, providing a means to adapt the model efficiently.

Assuming the parameters of the pretrained model are denoted as $\theta$, where the addition operation is defined as $\oplus$ and the negation operation as $\ominus$. For instance, for LoRA module, the computation for the intermediate layer weight matrix $\textbf{\textit{w}}$ is expressed as $\textbf{\textit{w}}^{\prime}=\textbf{\textit{w}}+\textbf{\textit{BA}}$, where $\textbf{\textit{B}}$ and $\textbf{\textit{A}}$ represent the two low-rank matrices in LoRA. Correspondingly, the activation values for this portion are $\textbf{\textit{x}}^{\prime}=\textbf{\textit{W}}\cdot \textbf{\textit{x}}+\textbf{\textit{BA}}\cdot \textbf{\textit{x}}$. To induce the LoRA module to forget certain knowledge, it suffices to take the negation of the activation change part, i.e., $\textbf{\textit{-BA}}\cdot \textbf{\textit{x}}$. The corresponding arithmetic operation is then:
\[
\ominus\theta_{lora}^{negation}=\ominus\theta_{lora}=\{\textbf{\textit{A}},-\textbf{\textit{B}}\}
\]

The above operation results in the reversal of activation values. The principle of this method is similar to gradient ascent, inducing a change in the intermediate layer's activation values in the direction opposite to gradient descent. This mechanism facilitates the unlearning of knowledge within the module.

\subsection{In-context learning}
Methods based on in-context learning differs from the previous two methods by not focusing on parameter operations. Instead, it views the model as a black box and utilizes its inherent in-context learning capability to supplement existing knowledge during inference. Despite consuming fewer resources and being relatively easy to execute, in-context learning methods modify the model's output results in an assisted manner with external knowledge during the question-and-answer process. As a result, it does not fundamentally erase harmful knowledge stored internally in the model.

\textbf{ICUL (arXiv 2023.10)} This paper introduces the novel in-context learning based unlearning method, applicable in scenarios where access is limited to the model's API without visibility into its internal parameters. In the context of inference using input prompts, ICUL leverages unlearning samples like $input_f$ and their corresponding answers as prompt examples. These examples aim to rectify outputs through prompt-based learning, relying on a limited set of samples. Specifically, for an unlearning sample, $input_f$, the following three steps are undertaken:

1) Flip the label of unlearning sample. Flip the label of the unlearning sample $input_f$ to $label_f$, obtaining the flipped sample ($input_f$, $label_f$).

2) Prompt Examples. The $s$ normal samples combined with the above unlearning sample is used to form the final prompt. $\left(input_f,label_f\right)\setminus$
$n\left(input_1,label_1\right)\setminus$$n\left(input_2,label_2\right)\setminus$$n...\setminus$
$n\left(input_s,label_s\right)$

3) Inference. The above demonstrations are used with the query $input_q$ to form the final prompt:

$\left(input_f,label_f\right)$

$\left(input_1,label_1\right)$
 
$\left(input_2,label_2\right)$

…

$\left(input_s,label_s\right)$

$input_q:$

While ICUL abstains from the need for modifying the model's parameters and treats the model as a black box, its limitation stems from reliance on the model's capability of in-context learning. If the model demonstrates a weak aptitude for in-context learning, its ability to glean insights from the provided examples and adapt its output is consequently diminished. In addition, when there exists a considerable semantic gap between $input_q$ and $input_f$, as in the case of multi-hop question answering, the effectiveness of this method in unlearning may not be optimal.

\begin{table*}[]
\caption{Statistics of the evaluation datasets.}
\resizebox{\linewidth}{!}{
\begin{tabular}{c|cccc}
\hline
\multirow{2}{*}{\textbf{Dataset}} &
  \textbf{Num. of sample} &
  \multirow{2}{*}{\textbf{Task}} &
  \multirow{2}{*}{\textbf{Application}} &
  \multirow{2}{*}{\textbf{\begin{tabular}[c]{@{}c@{}}Related\\ works\end{tabular}}} \\
                                           & \textbf{Train/Dev/Test} &                &                                &          \\ \hline
Part of Training Data Extraction Challenge & 15000                   & generation     & privacy information protection & KUL      \\ \hline
IMDB                                       & 20000/2000/25000        & classification & privacy information protection & EUL      \\ \hline
SAMSum                                     & 14732/818/819           & generation     & privacy information protection & EUL      \\ \hline
Part of PKU-SafeRLHF + TruthfulQA          & --                      & generation     & unlearning harmful content     & LLMU     \\ \hline
Harry Potter + BookCorpus                  & --                      & generation     & copyright content protection   & LLMU     \\ \hline
Part of HaluEval + TruthfulQA              & --                      & generation     & unlearning model hallucination & LLMU     \\ \hline
Part of Enron                              & 25000                   & classification & privacy information protection & DEPN     \\ \hline
Part of Civil Comments                     & 1000                    & classification & unlearning harmful content     & CPEM, TA \\ \hline
\end{tabular}
}
\end{table*}

\section{Dataset \& Evaluation}
This section introduces the datasets and their usage in evaluating the unlearning effects of LLMs. From the perspective of datasets and tasks, we categorize these datasets into classification and generation datasets. Details of these datasets are listed in Table 2.

\subsection{Classification Datasets}
\textbf{IMDB Dataset} The IMDB dataset is a sentiment classification dataset that includes user reviews of movies, directors, and actors, etc. It is divided into two sentiment categories. Both the training and testing sets contain 25,000 samples each. During testing, the particular movie or person names are randomly selected, and the trained model is tasked with forgetting all comments related to these specific movies or person from the training set.

\textbf{Enron Dataset} Enron derives from the emails and related documents of the Enron corporation, comprising approximately 500,000 emails involving communication among thousands of employees. This dataset has widespread applications in research areas such as fraud detection, social network analysis, and privacy protection. DEPN sampled 25,000 examples from the dataset of 500,000 for evaluating privacy information unlearning.

\textbf{Civil Comments Dataset} Civil Comments consists of public comments from news websites. In the complete dataset, the training/dev/testing sets are comprised of 1,804,874/9,732/97,320 samples, respectively. Each sample (comment) is labeled with a toxicity tag indicating the degree of toxicity in the comment. When using this dataset, CPEM and TA select a subset of samples with a high toxicity score of 0.8 as the training set to obtain a model targeting toxicity. During testing, a proposed unlearning method is employed to mitigate the toxicity of the post-training model and assess the normalized language expression of the detoxified model.

\subsection{Generation Datasets}
\textbf{SAMSum Dataset} SAMSum is a dataset designed for generative dialogue summarization, commonly employed in summarization tasks. The dataset consists of 14,732/818/819 samples in the training, development, and test sets, respectively. During testing, a specific speaker is randomly selected, and the model is tasked with forgetting all dialogues containing information about that selected speaker.

\section{Conclusion \& Future Directions}
Knowledge unlearning is a technique used to refine the internal storage of knowledge in large language models, aiming to prevent the generation of harmful information during use and safeguard ordinary users from potential harm. This paper provides an overview of research on knowledge unlearning for LLMs, categorizes existing methods, explains their principles and characteristics, and summarizes evaluation datasets and tasks employed in existing studies. Finally, challenges and future prospects in this domain is analyzed. Our goal is to encourage further research in this area and promote the responsible development of LLMs.

Although the knowledge unlearning technique for large language models has great potential, it is still in the early stage of exploration and there is insufficient research. For example: 
\begin{itemize}
\item There is a risk of catastrophic unlearning, especially in scenarios involving continuous unlearning request. Since most methods have been tested only on a limited retention set, while this small-scale test set indicates that the model has not experienced catastrophic forgetting, it may still lead to untested knowledge deficiencies. 

\item The cross-lingual and cross-modal generalization of forgetting methods is important. For multilingual models like mBERT \cite{devlin2019bert}, GPT-4 \cite{openai2023gpt4}, and LLaMA-2 \cite{touvron2023llama2}, which store cross-lingual knowledge of hundreds of languages and share a unified representation space among different languages, it is necessary to consider whether the knowledge unlearning for one language (such as English) will generalize to other languages (such as German, French, and Chinese). Similarly, for multimodal models like CLIP \cite{radford2021learning}, BLIP-2 \cite{li2023blip2}, and GPT-4 \cite{openai2023gpt4}, the cross-modal effects of forgetting methods need to be considered. 

\item Forgetting knowledge on specific topics while ensuring fundamental language understanding within that topic remains intact is a critical focus for future research.
\end{itemize}

In addition, our survey results show that current research on knowledge unlearning is primarily focused on pre-trained language models. However, for open-domain question answering LLMs like ChatGPT \cite{openai2022chatgpt}, GPT-4 \cite{openai2023gpt4}, LLAMA-2 \cite{touvron2023llama}, and BaiChuan \cite{yang2023baichuan}, relevant unlearning methods need to consider parameter scale and evaluation data. Currently, only LLMU and AU have conducted exploratory studies using gradient-based methods. On the other hand, for closed-source models like ChatGPT and GPT-4, especially considering the black-box nature under restricted access conditions where only the input and output of the model are accessible, it is important to focus on knowledge unlearning methods for these black-box models.

\bibliographystyle{unsrt}
\bibliography{abs}

\end{document}